\documentclass[11pt,letterpaper]{article}

\usepackage[T1]{fontenc}
\usepackage[utf8]{inputenc}
\usepackage{amsmath,amssymb}
\usepackage{lmodern}
\usepackage[margin=1in]{geometry}
\usepackage{microtype}
\usepackage{graphicx}
\usepackage{booktabs}
\usepackage{array}
\usepackage[labelfont=bf,labelsep=period,font=small,justification=justified,singlelinecheck=false,skip=5pt]{caption}

\usepackage[numbers,square,sort&compress]{natbib}
\usepackage{xcolor}
\definecolor{linkblue}{HTML}{1F4E99}
\usepackage{xurl}
\usepackage[colorlinks=true,linkcolor=linkblue,citecolor=linkblue,urlcolor=linkblue]{hyperref}
\usepackage{doi} 
\usepackage{fancyhdr}
\fancypagestyle{main}{\fancyhf{}\fancyfoot[C]{\thepage}}
\fancypagestyle{plain}{\fancyhf{}\fancyfoot[L]{\small Preprint.}\fancyfoot[C]{\thepage}}  
\title{Scaling large language model-assisted line-of-therapy adjudication with human-in-the-loop evidence-calibrated selection}

\author{%
  Olatunji Oni\thanks{Corresponding author: \href{mailto:oni.oa7@gmail.com}{oni.oa7@gmail.com}} ,
  Sanhita Sengupta, Marc De Benedetti, and Nathanial Eddy\\[4pt]
  Bristol Myers Squibb, NJ, USA}

\date{}

\begin{document}

\maketitle

\begin{abstract}
Large language models (LLMs) have enabled the ability to scale clinical adjudication tasks traditionally dependent on exhaustive manual review. Line-of-therapy (LoT) adjudication, a major bottleneck in oncology cohort construction, is a prime example. Line boundaries must be inferred from longitudinal treatment histories according to nuanced, study-specific definitions, at every regimen transition for every patient. LLMs can apply this adjudication at scale, but their reliability varies across patients. Treatment histories differ in complexity, and ambiguous or atypical patterns depend on expert clinical judgment. Reliable automation thus requires verified support for model assignments and the ability to identify cases that warrant manual review. Here, we present evidence-calibrated selection (ECS), a human-in-the-loop framework for scaling LLM-assisted LoT adjudication. ECS derives confidence from observed clinician adjudications, prioritizes patients with the least-supported assignments for follow-up review, and returns completed adjudications to the model as precedent. As the evidence base grows, ECS continuously calibrates confidence to estimate residual error, thus enabling more efficient use of clinician time. In simulations on 136 clinician adjudicated patients with multiple myeloma, ECS achieved a residual error of 5\% after review of 44.6\% of patients, compared to 62.2\% when review was directed by LLM self-reported confidence and 81.9\% under random sampling. These results support a scalable approach to study-specific LoT adjudication, using clinician evidence to inform automated predictions, minimize error, and reduce review burden.
\end{abstract}

\section{Introduction}

Large language models (LLMs) possess exceptional capability in extracting and interpreting clinical information with little or no task specific training \citep{agrawal2022clinical,huang2024chatgpt,moulson2024egfr}. This has made it possible to scale clinical adjudication tasks that have traditionally relied on exhaustive manual review. Line-of-therapy (LoT) adjudication is an especially challenging and high value objective for automation of this kind. In oncology, LoT defines a patient's particular stage of treatment and is a key element in cohort construction and outcome estimation. In these applications, approaches like feasibility assessments, real-world outcome benchmarks and external or contextual comparator cohorts are typically defined by therapy lines \citep{burcu2020external,fda2023externally}. The effect of treatment can vary by LoT so an incorrect assignment can have a significant impact on cohort composition and additional downstream efficacy estimates.

Despite this essential function, accurately assigning LoT remains challenging as it must be inferred from longitudinal medical records, and interpreted according to nuanced, study-specific definitions \citep{saini2021lines}. This is further complicated by the observation that clinical intent is rarely documented in real-world data sources. Especially illustrated across oncology, patients can receive multi-drug combinations over various lines, with induction, transplant, consolidation, and maintenance phases that do not necessarily constitute new lines \citep{rajkumar2015guidelines,kumar2017myeloma}. Over the course of treatment, therapeutic agents are added, dropped, or substituted, treatment gaps occur, and regimens overlap or restart \citep{meng2019temporal}. These changes can indicate a variety of clinical intents, including responses to toxicity, therapeutic maintenance, or a switch at progression, each of which carries a different line assignment. Furthermore, classification rules are inherently study dependent. For example, therapeutic transitions such as an intra-class substitution, continuation, or prior perioperative regimen may be considered a new line only under specific study questions \citep{fda2023externally,saini2021lines}. While manual adjudication allows clinicians to accurately resolve these considerations with respect to the study protocol, this process is resource-intensive, creating a substantial bottleneck for larger studies.

LLMs offer a practical advantage here as they can be given the LoT definition of a study and used to adjudicate patients at scale. However, patient treatment histories and model assessments are not uniform in complexity. Medical records that follow a typical course can be easily resolved by the predefined definitions, but ambiguous or atypical treatment patterns depend on expert review. LLM assignments thus require oversight, and the problem becomes how to best allocate limited expert review, deciding which patients clinicians should adjudicate and how to quantify and minimize error. This is similar to an active learning setting where we want to select the samples our model is least confident in to label \citep{settles2009active,lewis1994sequential}. Unfortunately, the most readily available measure of LLM certainty, the model's self-reported confidence, is often poorly calibrated, particularly for tasks that require specialized knowledge \citep{xiong2024uncertainty,savage2025uncertainty}, and therefore offers a weak basis for allocating review and for judging total error. Reliable automation thus requires a confidence grounded in empirical support from clinician-validated observations, as well as a mechanism for returning this context to the model so that each review can inform subsequent predictions.

Here, we present evidence-calibrated selection (ECS), a human-in-the-loop framework for scaling LLM-assisted LoT adjudication (Figure 1). ECS derives the confidence of each LLM assignment from the empirical support provided by clinician adjudications as they accumulate during review. Patients with the least supported assignments are prioritized for adjudication, and the completed reviews are then added to the evidence base and used to evaluate subsequent assignments in order to provide the LLM with validated precedent. As this evidence grows, the resulting confidence also provides an estimate of residual error among remaining patients, which can be used to guide further review and concentrate clinician effort. And because the evidence is generated under the study's own LoT definition, the framework organically adapts to the clinical interpretation of the study. To evaluate this approach, we simulate review using clinician adjudicated treatment histories from patients with multiple myeloma, and compare the performance of ECS to random review, review guided by LLM self-reported confidence and an oracle selection strategy.

\begin{figure}[!htb]
  \centering
  \includegraphics[width=1.000\textwidth]{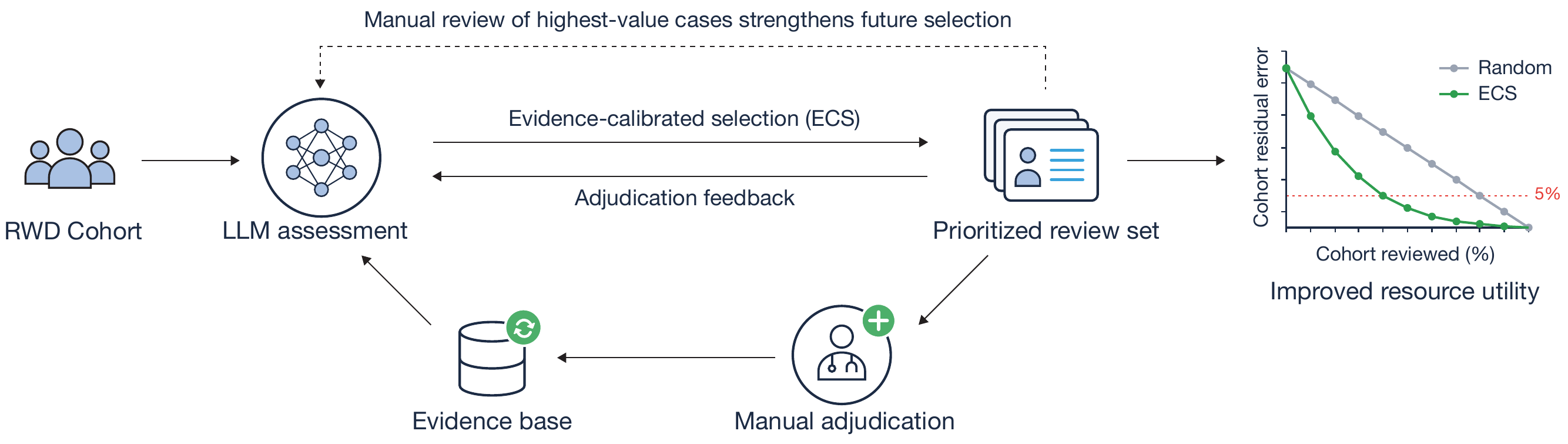}
  \caption{\textbf{Overview of evidence-calibrated selection.} LLM assigns a LoT across the treatment histories of a real world cohort. ECS estimates confidence of each assignment based on adjudications already collected by clinicians on similar decisions and prioritizes the patients with the least supported assignments for manual adjudication. Each completed review is added to the evidence base used to score the next assignments, and is fed back to the model as a precedent. The right schematic shows the resulting reduction in residual error relative to random review.}
  \label{fig:1}
\end{figure}

\section{Methods}

\subsection{Evidence-calibrated confidence}

ECS estimates the confidence of each LLM call based on the empirical frequency of clinician adjudications currently seen for comparable regimen transitions. For each transition from one regimen to the next the LLM makes a binary call as to whether the line advances. Given a manual review budget, clinicians can directly confirm or reverse the LLM calls providing a verified evidence base. Following an empirical-Bayes approach \citep{efron1975stein}, the observed rate of line advancement within a transition pattern informs a prior for the error probability of each LLM call type. This prior is then updated with the observed frequency of clinician reversal of calls of that type within a pattern under a Beta--binomial model \citep{gelman2013bda}. Confidence thus reflects the amount of evidence available and is driven primarily by the pattern level prior when there are few adjudications and gradually converges to the observed accuracy of the calls of that type as adjudications accumulate. As such, the confidence is a measure of the reliability of the model against the study's own reference standard and becomes increasingly precise as the review progresses.

From the treatment history of each patient, each transition, representing the observable change between consecutive regimens, is grouped with similar transitions, based on whether the line-defining agents are switched, added, removed, or repeated. Any supportive agents such as corticosteroids are excluded from this grouping. In addition, transition patterns can be refined by standard LoT descriptions, such as progression or long treatment gaps over 180 days, provided the adjudicated transitions show that the description changes a clinical decision. To investigate this, we use a Bayes factor threshold of 3 \citep{kass1995bayes}. Using the adjudicated transitions in pattern $g$, the consistency of clinician decisions for a transition $t$ is given as
\begin{equation}
  r_g = \frac{a_g + 1}{n_g + 2}
\end{equation}
\noindent where $n_{g}$ is the number of adjudicated transitions in the pattern, and $a_{g}$ is the number adjudicated as advances. Note that if a pattern is almost always advanced by clinicians or virtually never advanced, then its consistency is close to one or zero, a contested pattern has a consistency close to a half. So, the prior probability of an LLM call being overturned, $\pi_{t}$, is $r_{g}$ if the model makes no advance, and $1-r_{g}$ if it does. We then extract evidence about the particular decision from the stratum $s$ defined by the pattern and LLM call type, and count $e_{s}$ of the $n_{s}$ adjudicated calls in this stratum as overturned. In a Beta--binomial model, the prior contributes $\kappa$ pseudo-observations \citep{gelman2013bda} and the probability that the call is inaccurate is calculated by the posterior mean
\begin{equation}
  \hat{p}_t = \frac{e_s + \kappa\,\pi_t}{n_s + \kappa}, \qquad c_t = 1 - \hat{p}_t
\end{equation}
\noindent with $\kappa=4$. Accordingly, when few comparable calls have been adjudicated, the estimate remains close to the prior from the pattern, but converges to the observed rate at which clinicians overturn calls of that kind as adjudications accumulate. A precedent can also change a call, in which case the new call is wrong exactly when the old call was correct, and its confidence is $\hat{p}_t$. The confidence that every call for patient $i$ is correct is then
\begin{equation}
  C_i = \prod_{t \in i} c_t
\end{equation}
\noindent where the product is taken over all calls for patient $i$ with each call treated as independent. This allows each call to be scored according to how consistently clinicians resolve a given transition pattern and how often calls of that pattern have been overturned. Calls that are in line with how clinicians consistently resolve a particular pattern of transition will rapidly gain confidence as the review progresses. Calls that are in conflict with the established consensus, or which fall into a rare or contested pattern, will remain at low confidence until there are sufficient comparable calls to adjudicate and support them. Patients can thus be ranked by their confidence with the lowest confidence being assigned to patients whose treatment histories contain poorly supported transitions. The confidence of each patient is a single value which can be used to both prioritize the review and to estimate the error remaining amongst the unreviewed patients.

\subsection{Selection and error quantification}

In each round of review, clinicians adjudicate the $k$ patients with the smallest $C_{i}$ that have not yet been reviewed. Each transition is assigned a confidence of 0.5 before any adjudication. The first round then organically selects patients with the longest treatment histories. So there is no need to set any random seed. Moreover, since $C_{i}$ is a calibrated probability, the residual error of the cohort can be directly estimated from the confidence of the patients not reviewed yet as
\begin{equation}
  \varepsilon = \frac{1}{N} \sum_{i \in U} \left( 1 - C_i \right)
\end{equation}
\noindent where $U$ is the set of unreviewed patients, and $N$ is the size of the cohort. Since $\varepsilon$ is updated with every round of adjudications added to the evidence base, we get a running estimate of the error expected to remain in the cohort during review. Precedent is also added to the LLM prompt as rules from adjudicated transitions after each round.

\subsection{Dataset}

Treatment histories were obtained from the COTA Vantage multiple myeloma (MM) database (Verana Health), a de-identified, longitudinal, electronic health record (EHR)-derived data set licensed by Bristol Myers Squibb (BMS). The database is a curated data set from a US provider network of academic, community, and hospital system sites. Structured EHR fields are programmatically ingested and augmented by manual abstraction of the clinical notes by trained abstractors into a version controlled, disease specific data model. Data is provided with vendor quality assurance in place including programmatic plausibility checks, record audits and cross functional review of each data delivery. The records cover the period from diagnosis to the last recorded encounter. Systemic therapy is recorded as individual regimen records, and a new regimen record is created when an antineoplastic agent is added or removed, or when a documented treatment hold is greater than 30 days. The vendor’s standard curation excludes patients diagnosed before the age of 18 and patients with a concomitant primary malignancy. The study cohort comprises 136 patients with MM who met prespecified eligibility criteria for an external comparator analysis, identified by BMS through vendor patient identifiers obtained in Q2 2025, with LoT assignments clinically adjudicated for a prior study. Data are de-identified in accordance with the Health Insurance Portability and Accountability Act (HIPAA) Privacy Rule.

\subsection{LLM configuration and prompting}

We used GPT-5-mini \citep{openai2025gpt5}, a cost-efficient reasoning model designed for well-defined tasks with precisely specified instructions. The specific model version \texttt{gpt-5-mini-2025-08-07} was accessed using the Azure OpenAI API with a deployment in the East US 2 region. The inference was done in August–September 2026 with medium reasoning effort and a limit of 4,000 completion tokens, including both reasoning tokens and visible-output tokens. The treatment histories of patients were processed separately. Each request contained the regimen history in chronological order, along with the therapies that had been given, the dates of the treatments and the clinical context that was available.

The prompt specified the adjudication task, ordered decision rules, confidence assessment, and required output format. The model was instructed to exclude supportive and adjunctive agents from regimen comparisons, assign the first regimen to LoT 1, and evaluate each subsequent regimen relative to its immediate predecessor. Decision rules were applied in a fixed order, with the first applicable rule determining the transition. The current LoT was maintained when the subsequent regimen met the first applicable condition among (1) planned consolidation without intervening progression or relapse, (2) continuation of unchanged line-defining therapies irrespective of treatment gap or response, and (3) single-agent step-down maintenance without intervening progression or relapse. Otherwise, LoT advanced following additions, substitutions, switches, or removals of line-defining therapies. Advancing transitions were further classified according to whether progression or relapse was documented. During iterative review, completed clinician adjudications were incorporated into subsequent prompts as contextual precedents for LoT assignment.

The model was instructed to return a JSON array, with one object per input regimen in the original order. For each transition, the requested fields comprised a binary advancement indicator, the cumulative LoT number, a transition category, a one-sentence rationale citing supporting evidence and qualifying any inference, and a self-reported confidence score from 0 to 10 expressing the estimated likelihood that a clinician adjudicator would agree with the advancement decision. LoT increased by one for each advancing transition and otherwise remained unchanged.

\subsection{Simulation and evaluation}

The three review strategies compared with ECS include random review, uncertainty sampling on LLM self-reported confidence \citep{lewis1994sequential}, and an oracle that reviews the patients whose current assignments are wrong. In uncertainty sampling, we consider patient confidence to be the product of the model's transition confidences, rescaled to the unit interval. The objective of this comparison was to isolate the effect of the selection rule. To do this, all strategies received the same precedent updates and reviewed the same number of patients per round. In the primary analysis, 100 sub-cohorts of 100 patients were used to compare strategies, each sampled without replacement from the full cohort, and seven patients were reviewed per round. To express effort in review rounds, the strategies were repeated with five patients per round, and across batch sizes of 2 to 20 patients per round. Each strategy was also applied to the full cohort of 136 patients. A patient was deemed correct when every transition matched the clinician, and reviewed patients took the clinician's assignment. The performance of each strategy was evaluated by accuracy among unreviewed patients, residual error, and review burden. Here, residual error refers to the number of unreviewed patients with at least one incorrect call divided by $N$, and review burden refers to the share of the cohort reviewed before residual error fell to 5\% and remained there. Using the expected calibration error (ECE) over ten equal-width bins \citep{naeini2015bbq,guo2017calibration} and the Brier score \citep{brier1950verification}, we assessed the calibration of patient confidence among unreviewed patients at every round after the first. The estimated residual error was compared with the observed residual error at every round, and confidence intervals (CIs) were obtained from 2,000 bootstrap resamples of sub-cohorts.

\section{Results and discussion}

\subsection{LLM errors concentrate across transition clinical intent}

Before examining the different review strategies, we investigated where the LLM made its mistakes throughout the treatment histories of the entire cohort. By grouping the transitions according to the change in line-defining agents and, within each type, according to the drug classes concerned, it was found that the errors were concentrated in a small number of transition patterns (Table 1). Transitions to a regimen based on a different drug class accounted for 62.6\% of all transitions and were classified as new lines in 99.4\% of instances, yet they represented only 5.4\% of the LLM errors. These types of transition follow the normal course of treatment and so the LoT definition can deal with them with minimal need for interpretation. Conversely, transitions involving high-dose melphalan and autologous transplant behaved differently. Whether the transplant was regarded as a new regimen, was added to an existing one, or was removed from it, these transitions made up 8.7\% of all transitions but amounted to 34\% of the LLM errors. The line assignments for these cases were dependent on whether the transplant was given to consolidate a planned induction or if it was administered at relapse \citep{rajkumar2015guidelines}, a distinction which is frequently not explicitly stated in the records. In these cases, clinicians decided to advance the line in between 80\% and 84\% of the cases. Overall, the most disputed category of transitions was that of repeated regimens, with only 27.1\% of them being judged as new lines. Among this group, a repeated proteasome inhibitor and immunomodulatory doublet, for example bortezomib and lenalidomide, was advanced in 47.1\% of cases, and its 17 transitions accounted for 14.3\% of all errors. Likewise, the handling of removals depended on the drug classes involved, with the removal of an immunomodulatory drug being advanced in 33.3\% of cases and the removal of transplant components in 81.3\%.

\begin{table}[!htb]
  \caption{\textbf{Clinician advance rates and LLM errors by transition type and drug class. }Transitions in the full cohort grouped by the change in line-defining agents between consecutive regimens, with supportive agents excluded, and within each type by the drug classes involved. Clinician advance rate indicates the share of transitions adjudicated as a new LoT. LLM errors indicate calls made without adjudicated examples that disagreed with the clinician, shown as counts and as a share of all 56 errors.}
  \label{tab:1}
  \footnotesize
  \renewcommand{\arraystretch}{1.1}
  \begin{tabular}{@{}p{3.15in}>{\centering\arraybackslash}p{0.95in}>{\centering\arraybackslash}p{0.95in}>{\centering\arraybackslash}p{0.95in}@{}}
    \toprule
    \textbf{Transition type and drug classes involved} & \textbf{Transitions, n (\%)} & \textbf{Clinician advance rate (\%)} & \textbf{LLM errors, n (\%)} \\
    \midrule
    \textbf{Switch} & \textbf{587 (75.3)} & \textbf{98.0} & \textbf{13 (23.2)} \\
    \hspace*{12pt}To a different drug class & 488 (62.6) & 99.4 & 3 (5.4) \\
    \hspace*{12pt}Into high-dose melphalan and autologous transplant & 37 (4.7) & 83.8 & 7 (12.5) \\
    \hspace*{12pt}Within the same drug classes & 34 (4.4) & 94.1 & 2 (3.6) \\
    \hspace*{12pt}Into BCMA-directed or cellular therapy & 28 (3.6) & 96.4 & 1 (1.8) \\
    \textbf{Addition} & \textbf{110 (14.1)} & \textbf{87.3} & \textbf{17 (30.4)} \\
    \hspace*{12pt}Immunomodulatory drug added & 38 (4.9) & 81.6 & 7 (12.5) \\
    \hspace*{12pt}Anti-CD38 antibody added & 21 (2.7) & 95.2 & 1 (1.8) \\
    \hspace*{12pt}Proteasome inhibitor added & 19 (2.4) & 94.7 & 1 (1.8) \\
    \hspace*{12pt}Other agent added & 17 (2.2) & 88.2 & 2 (3.6) \\
    \hspace*{12pt}High-dose melphalan and autologous transplant added & 15 (1.9) & 80.0 & 6 (10.7) \\
    \textbf{Repeat} & \textbf{48 (6.2)} & \textbf{27.1} & \textbf{13 (23.2)} \\
    \hspace*{12pt}Proteasome inhibitor and immunomodulatory doublet & 17 (2.2) & 47.1 & 8 (14.3) \\
    \hspace*{12pt}Anti-CD38 combination & 15 (1.9) & 13.3 & 3 (5.4) \\
    \hspace*{12pt}Single agent & 11 (1.4) & 18.2 & 1 (1.8) \\
    \hspace*{12pt}Other combination & 5 (0.6) & 20.0 & 1 (1.8) \\
    \textbf{Removal} & \textbf{35 (4.5)} & \textbf{62.9} & \textbf{13 (23.2)} \\
    \hspace*{12pt}Transplant components removed & 16 (2.1) & 81.3 & 6 (10.7) \\
    \hspace*{12pt}Immunomodulatory drug removed & 9 (1.2) & 33.3 & 4 (7.1) \\
    \hspace*{12pt}Proteasome inhibitor removed & 8 (1.0) & 50.0 & 2 (3.6) \\
    \hspace*{12pt}Other agent removed & 2 (0.3) & 100.0 & 1 (1.8) \\
    \midrule
    \textbf{All transitions} & \textbf{780 (100.0)} & \textbf{90.5} & \textbf{56 (100.0)} \\
    \bottomrule
  \end{tabular}
\end{table}

Accordingly, the same observed change, for example a doublet that pauses and then resumes without change, can be interpreted as either continuation after a treatment holiday or re-treatment at progression. Typically, clinicians solve these cases by using contextual information that is not explicitly documented in the record. The LLM does well at applying the written definition for typical treatment courses, but its mistakes are concentrated in these less common transitions, whose interpretation depends on the drug classes involved. Hence, a confidence measure should encode the amount of clinician evidence for each type of transition. In ECS, the transition patterns are built to cluster each decision with similar adjudicated cases, without requiring prior specification of the patterns in Table 1.

\subsection{ECS reaches 5\% residual error with less clinician review}

When evaluated with 100 sub-cohorts, ECS resulted in a  faster decrease in residual error, compared with both random review and review based on LLM confidence (Figure 2A). Specifically, ECS reduced residual error to 5\% after reviewing 44.6\% of the cohort (95\% CI 43.5--45.7), whereas LLM confidence and random review required 62.2\% and 81.9\% of the cohort, respectively. The oracle required just 29.8\% of the cohort (Table 2). Within the same sub-cohorts, direct comparison of the two methods showed that ECS resulted in a 28\% relative reduction in the review burden compared to LLM confidence and was consistently less review-intensive across all 100 sub-cohorts. This trend was also observed when analyzing the errors of the patients that have not been reviewed yet (Figure 2B). After reviewing 28\% of the cohort, 85.4\% of the remaining patients were fully correct under ECS, versus 77.7\% under LLM confidence and 70.1\% under random review. This order was preserved in the entire cohort of 136 patients, as ECS achieved 5\% residual error after 63 patients (46\%) were reviewed, whereas LLM confidence and random review required on average 84 (62\%) and 112 (82\%) patients, respectively.

The observed improvements can be attributed to the targeted manual assessments. The LLM’s self-reported confidence scores were largely clustered within a narrow range, with 96\% of the transitions scoring 8 or higher out of 10. Thus, a patient’s level of confidence was largely a function of the length of their treatment history and so provided only a limited means of discriminating between different probabilities of error. Conversely, patients with rare or contested transitions in their medical histories expressed significantly lower confidence in ECS (Table 1). In case of patients not reviewed and with confidence below 0.5, 62\% had at least one wrong assignment according to ECS while only 24\% according to LLM confidence. ECS also selected groups of seven patients with an average of 2.0 patients with an incorrect assignment, while groups selected on the basis of LLM confidence contained 1.2.

\begin{figure}[!htb]
  \centering
  \includegraphics[width=1.000\textwidth]{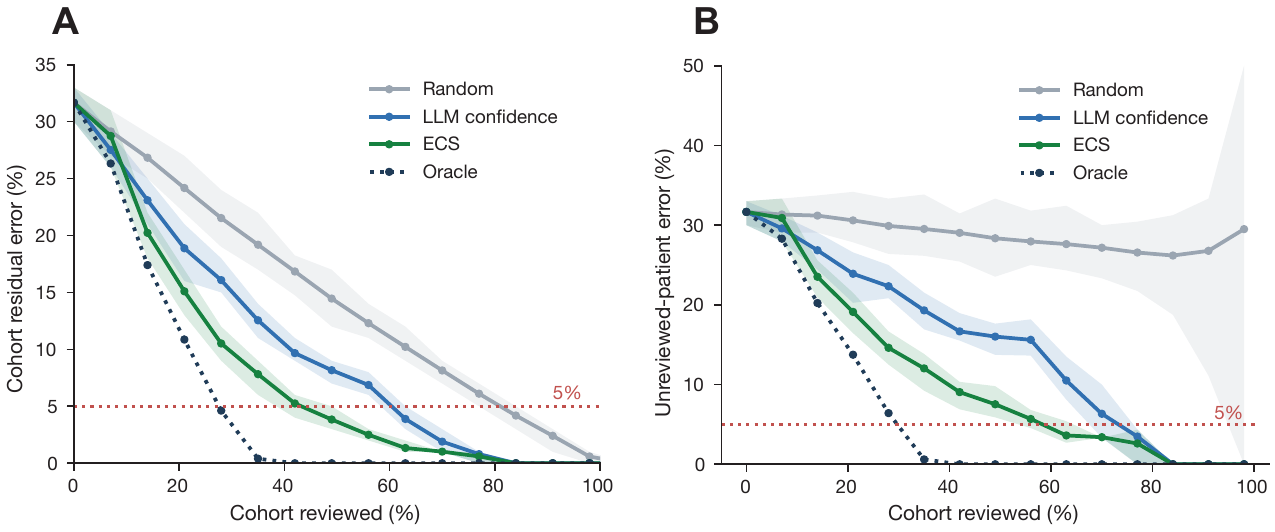}
  \caption{\textbf{ECS reduces residual error with less clinician review.} (\textbf{A}) Residual error across the cohort as a function of the share of patients reviewed, where reviewed patients are given the clinician's assignment. (\textbf{B}) Error among patients not yet reviewed. Random (gray) selects patients in random order. LLM confidence (blue) selects patients with the lowest LLM self-reported confidence first. ECS (green) selects patients with the lowest evidence-calibrated confidence first. Oracle (dotted navy) selects only patients whose current assignments are wrong, serving as an idealized benchmark for the best achievable selection. Results are shown for 100 sub-cohorts of 100 patients with seven patients reviewed per round. Lines indicate the mean, and shaded bands indicate the interquartile range. Red dotted line indicates 5\% residual error.}
  \label{fig:2}
\end{figure}

\begin{table}[!htb]
  \caption{\textbf{Review burden, review rounds, and accuracy by review strategy.} Values are means over 100 sub-cohorts of 100 patients. Review burden is the share of the cohort reviewed before residual error fell to 5\% and remained there, with seven patients reviewed per round. Confidence intervals (95\%) are from 2,000 bootstrap resamples. Rounds are from a separate simulation with five patients reviewed per round. Accuracy is the share of unreviewed patients with every transition correct after 14\%, 28\%, and 42\% of the cohort had been reviewed.}
  \label{tab:2}
  \footnotesize
  \renewcommand{\arraystretch}{1.1}
  \begin{tabular}{@{}p{1.0in}>{\centering\arraybackslash}p{1.27in}>{\centering\arraybackslash}p{1.2in}>{\centering\arraybackslash}p{0.732in}>{\centering\arraybackslash}p{0.732in}>{\centering\arraybackslash}p{0.732in}@{}}
    \toprule
    \textbf{Strategy} & \textbf{Review burden, \% (95\% CI)} & \textbf{Rounds to 5\%, mean (median)} & \textbf{Accuracy, 14\% (\%)} & \textbf{Accuracy, 28\% (\%)} & \textbf{Accuracy, 42\% (\%)} \\
    \midrule
    Random & 81.9 (80.5--83.2) & 16.3 (17) & 68.8 & 70.1 & 71.0 \\
    LLM confidence & 62.2 (61.5--63.0) & 12.2 (12) & 73.2 & 77.7 & 83.3 \\
    ECS & 44.6 (43.5--45.7) & 8.9 (9) & 76.5 & 85.4 & 90.9 \\
    Oracle & 29.8 (29.1--30.5) & 5.8 (6) & 79.8 & 93.6 & 100.0 \\
    \bottomrule
  \end{tabular}
\end{table}

\subsection{ECS requires fewer review rounds across batch sizes}

In practice, manual adjudication typically proceeds in rounds, where a batch of patients is reviewed by clinicians and the model is updated before the next batch is selected. The number of rounds is therefore a more direct measure of the time required to curate a cohort than the number of patients reviewed. With five patients reviewed per round, ECS reached 5\% residual error after 8.9 rounds on average, compared with 12.2 rounds for LLM confidence and 16.3 for random review (Figure 3A, B). ECS also required fewer rounds than LLM confidence in all 100 sub-cohorts. Since the absolute number of rounds depends on the choice of batch size, we repeated this comparison with 2 to 20 patients per round (Figure 3C). Although the number of rounds saved by ECS varied with the batch size, from 8.4 at two patients per round to 1.6 at ten, the relative reduction remained stable at 27--28\% for batches of up to 7\% of the cohort (Figure 3D). Due to less frequent updates of its estimates, and a larger share of patients in each batch that would not have been selected individually, the advantage of ECS narrowed to 19\% at 20 patients per round.

\begin{figure}[!htb]
  \centering
  \includegraphics[width=0.907\textwidth]{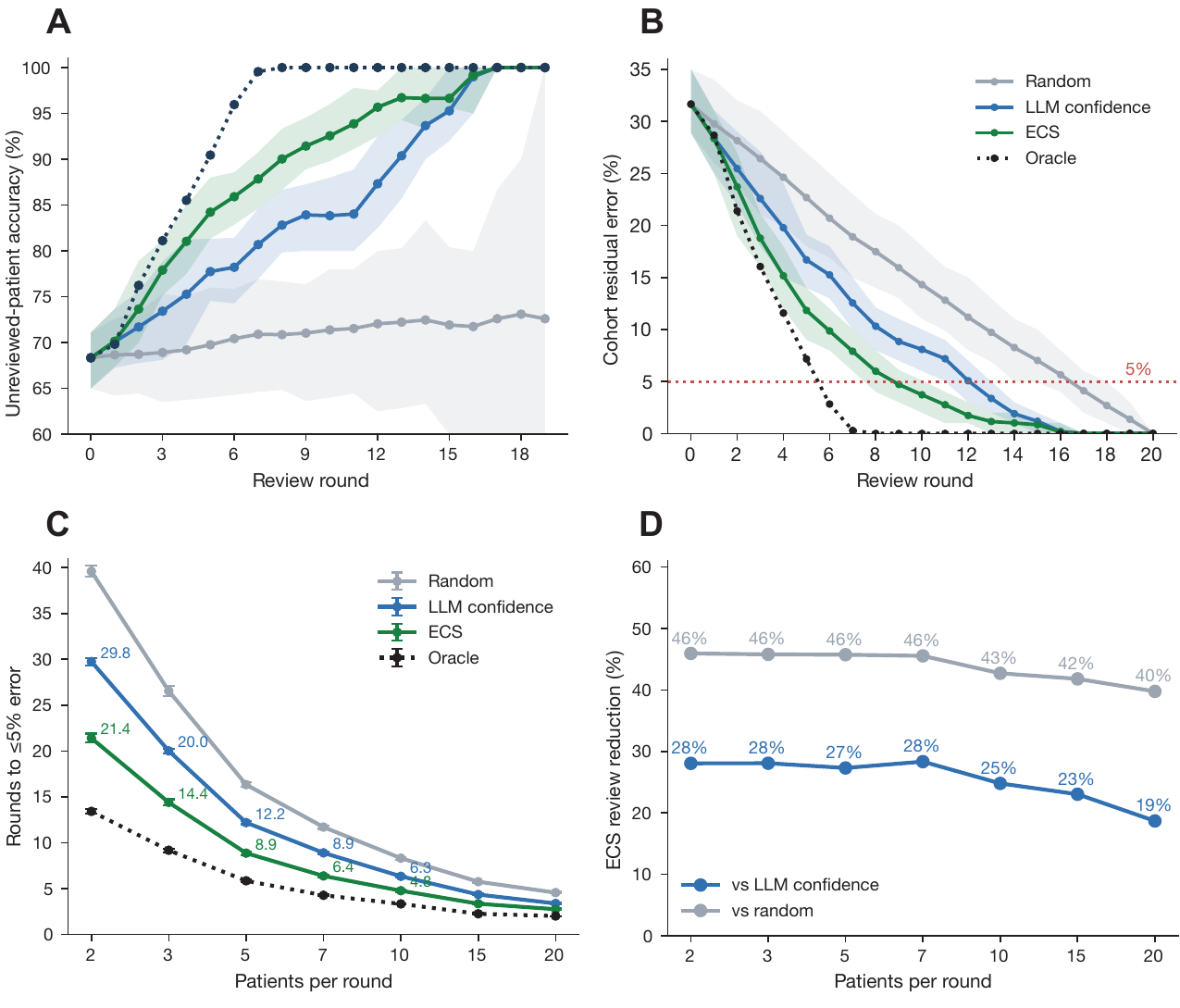}
  \caption{\textbf{ECS requires fewer review rounds.} (\textbf{A}) Accuracy among unreviewed patients and (\textbf{B}) residual error by review round, with five patients reviewed per round. Lines indicate the mean, and shaded bands indicate the interquartile range across 100 sub-cohorts. (\textbf{C}) Mean number of rounds required to reach 5\% residual error for batch sizes of 2 to 20 patients per round. Error bars indicate 95\% confidence intervals. (\textbf{D}) Relative reduction in review rounds achieved by ECS against LLM confidence and random review.}
  \label{fig:3}
\end{figure}

\subsection{ECS provides a calibrated estimate of residual error}

For individual patients, the ECS was well-calibrated with an ECE of 0.058 and Brier score of 0.093 (Figure 4A). Conversely, the LLM’s self-reported confidence, assessed on the same patients and calls, had an ECE of 0.278 and a Brier score of 0.195. Interestingly, for ECS, the deviations from perfect calibration were conservative, with patients with low confidence being correct more often than predicted. Calibrating confidence among patients not yet reviewed is required for the accurate measurement of residual error. As such, the residual error in a partially reviewed cohort cannot be directly observed. The estimated residual error in the ECS closely tracked the observed residual error over the review period (Figure 4B). The estimated residual error fell to 5\% as a median of 56\% of the cohort had been reviewed immediately after the observed residual error reached 5\% at 42\%. At this point, the residual error observed was 2.1\% and did not exceed 5\% in any of the sub-cohorts. Conversely, the estimates based on the LLM’s self-reported confidence consistently overestimated the residual error and did not fall below 5\% until 84\% of the cohort had been reviewed, 21 percentage points after the observed error (Figure 4C). Thus, uncalibrated confidence provides little information about the residual error and causes unnecessary review of patients that are already correctly assigned. However, a calibrated estimate does provide the ability to quantify the expected residual error in the unreviewed cohort at any point during the review process. This enables the scope of the review to be aligned with the error tolerance of a given study and enables the information to be reported alongside the population, enabling regulatory comparisons.

\begin{figure}[!htb]
  \centering
  \includegraphics[width=1.000\textwidth]{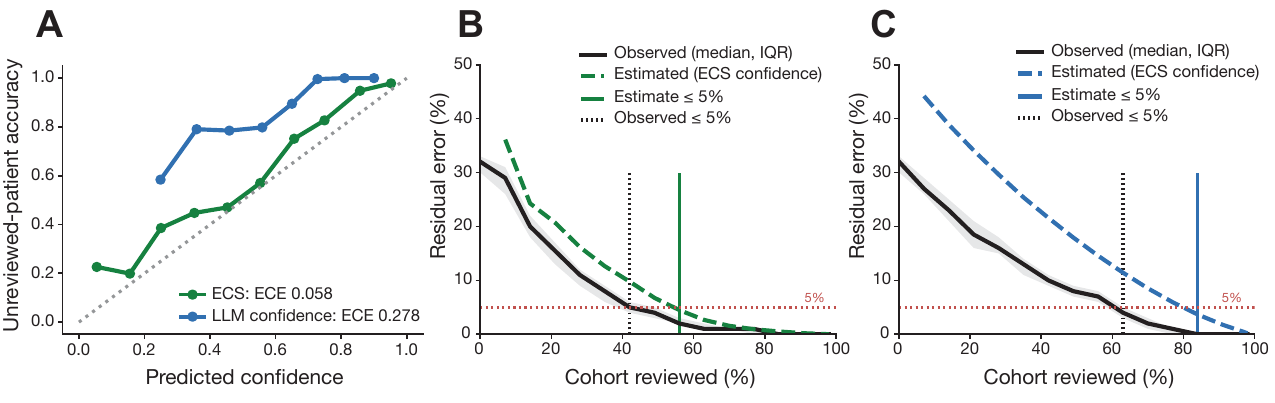}
  \caption{\textbf{ECS provides a calibrated estimate of residual error.} (\textbf{A}) Calibration of patient-level confidence for ECS and LLM self-reported confidence, evaluated on the same unreviewed patients at every round after the first (66,500 patient-rounds; bins with fewer than 200 patient-rounds are not shown). Dotted line indicates perfect calibration. (\textbf{B}, \textbf{C}) Observed residual error (median and interquartile range) and the residual error estimated from (\textbf{B}) ECS confidence in the ECS arm and (\textbf{C}) LLM self-reported confidence in the LLM-confidence arm. Vertical lines indicate where the estimated residual error first reaches 5\% (solid) and where the observed residual error first reaches 5\% (dotted).}
  \label{fig:4}
\end{figure}

\subsection{Scaling LoT adjudication in practice}

These results suggest that scaling LoT adjudication requires not only high overall accuracy but also a systematic assessment of model reliability at the level of individual assignments. LLM assignments are not equally reliable across patients because errors cluster in transitions, the interpretation of which depends on clinical intent and the study definition (Table 1). Furthermore, since ECS learns its confidence from the adjudications of the study itself, these patterns do not need to be specified a priori and the framework can adapt to whichever LoT definition the protocol applies. This approach is generalizable to other diseases, alternative definitions of lines, or eligibility criteria that require interpretation of patient records. In each new context, ECS identifies the transition patterns where errors are concentrated, rather than relying on predefined transitions. By shifting clinical review from exhaustive verification to targeted resolution of ambiguous transitions, ECS establishes a process in which each adjudication incrementally refines confidence estimates and informs subsequent LLM assignments. Notably, ECS operates without additional LLM queries, relying only on observed counts and a single prior, thereby introducing minimal overhead to existing adjudication workflows. The framework is also applicable to scoring assignments from rule-based algorithms, and supports selective classification by enabling high-confidence assignments to be accepted without further review using the same confidence thresholds \citep{geifman2017selective}.

\section{Conclusion}

We proposed evidence-calibrated selection, a human-in-the-loop framework for scaling LLM-assisted LoT adjudication. Since the reliability of LLM assignments will depend on the complexity of each treatment history, ECS estimates the confidence of predictions using the empirical frequency and concordance of observed physician adjudications across similar comparable cases. The confidence is used to target review on the patients with least support to their assignments and completed adjudications are given back to the model as precedent. In simulation, this lowered the amount of review needed to reach a threshold 5\% residual error by 28\% compared to review directed by LLM self-reported confidence, and by 46\% compared to random review. The reduction in review rounds was consistent across batch sizes, and the calibrated confidence gave an estimate of the error remaining among unreviewed patients throughout review. We propose a scalable approach to study-specific LoT adjudication that leverages physician evidence to drive automated predictions, minimize error, and reduce review load in constructing LoT-defined cohorts for external comparators, outcome benchmarks, and trial feasibility assessments.

\bibliographystyle{unsrtnat}
\bibliography{main}

\end{document}